\documentclass{article}

\usepackage[final]{neurips_2023}

\usepackage{paralist}
\usepackage{graphicx} 
\usepackage[utf8]{inputenc}
\usepackage[T1]{fontenc}
\usepackage{hyperref}
\usepackage{url}
\usepackage{booktabs}
\usepackage{amsfonts}
\usepackage{nicefrac}
\usepackage{microtype}
\usepackage{xcolor}
\usepackage{multirow}
\usepackage{amsmath}
\usepackage{physics}
\usepackage{todonotes}
\usepackage{wrapfig}
\usepackage{svg}

\title{Minimum Description Length Hopfield Networks}

\author{
Matan Abudy\textsuperscript{1}, Nur Lan\textsuperscript{1,2}, Emmanuel Chemla\textsuperscript{2}, Roni Katzir\textsuperscript{1}
\\\textsuperscript{1}Tel Aviv University, \textsuperscript{2}Ecole Normale Supérieure 
\\
\texttt{matan.abudy@gmail.com}
\\ \texttt{\{nur.lan,emmanuel.chemla\}@ens.psl.eu}
\\ \texttt{rkatzir@tauex.tau.ac.il}
}
  
\begin{document}

\maketitle

\begin{abstract}
Associative memory architectures are designed for memorization but also offer, through their retrieval method, a form of generalization to unseen inputs: stored memories can be seen as prototypes from this point of view. Focusing on Modern Hopfield Networks (MHN), we show that a large memorization capacity undermines the generalization opportunity. We offer a solution to better optimize this tradeoff. It relies on Minimum Description Length (MDL) to determine during training which memories to store, as well as how many of them.
\end{abstract}

\section{Introduction}
Generalization requires going beyond the training data. Hopfield Networks (HNs), both classical and modern, involve a retrieval component that can be seen as a form of generalization: an input image that is fed into a trained network can lead to a stored image even if the input itself is quite different from anything that the network saw during training. This form of generalization, however, is limited by the fact that the training stage of the network incentivizes rote memorization. 

The present paper explores the generalization capacity of HNs using a very simple setup. Our goals are twofold. We wish to both sharpen the implications of the memorization in HNs and offer a remedy that incorporates generalization into the storage phase. For our first goal, we note that rote memorization in the storage phase can lead to the storage of a wrong number of memories. For our second goal, we show how incorporating a well-studied principle of generalization — specifically, the principle of Minimum Description Length (MDL; \citealt{Solomonoff:1964,Rissanen:1978}) — into the storage phase can help address the problem that we note for rote memorization. MDL balances the complexity of a hypothesis against its fit to the data, and by doing so it leads to hypotheses that are relatively simple, and therefore generalize beyond the data, but not too simple.

We start, in section \ref{sec:bg}, with relevant background. We assume familiarity with HNs and focus on MDL and its use in the context of neural networks. Section \ref{sec:mdl:hn} presents MDL HNs. In section \ref{sec:exp:1} we illustrate the problem of incorrect number of memories, and how solving it can lead to good generalization. In section \ref{sec:exp:2} we show how MDL does this. Throughout, we will use modern HNs, but as far as we can tell our essential point holds in Classical HNs and other familiar architectures for associative memory. Section \ref{sec:disc} concludes and outlines remaining issues.

\section{\label{sec:bg}Background on simplicity and MDL}

Consider a hypothesis space $\mathcal{G}$ of possible networks, and a corpus of input data $D$. In our case, $\mathcal{G}$ is the set of all possible HNs expressible using our representations, and $D$ is a set of input images. For any $G\in \mathcal{G}$, we may consider the ways in which we can encode the data $D$, given that $G$. The MDL principle (\citealp{Rissanen:1978}), a computable approximation of Kolmogorov Complexity (\citealp{Solomonoff:1964,Kolmogorov:1965,Chaitin:1966}), aims at the $G$ that minimizes $|G|+|D:G|$, where $|G|$ is the size of $G$ and $|D:G|$ is the length of the shortest encoding of $D$ given $G$ (with both components typically measured in bits). Minimizing $|G|$ favors simple, general networks that often fit the data poorly. Minimizing $|D:G|$ favors complex, overly specific networks that overfit the data. By minimizing the sum, MDL aims at an intermediate level of generalization: reasonably simple networks that fit the data reasonably well. 

Simplicity criteria have been applied to  artificial neural networks, primarily for feed-forward architectures, as early as \citet{hinton_keeping_1993}, who minimize the encoding length of a network's weights alongside its error, and to \citet{ZhangMuhlenbein:1993,ZhangMuhlenbein:1995}, who use a simplicity metric that is essentially the same as the MDL metric that we use in the present work. \citet{Schmidhuber:1997} presents an algorithm for discovering networks that optimize a simplicity metric that is closely related to MDL. Simplicity criteria have been used in a range of works on neural networks, including recent contributions (e.g., \citealp{AhmadizarSoltanianAkhlaghianTabTsoulos:2015} 
and \citealp{GaierHa:2019}). More recently, \citet{LanGeyerChemlaKatzir:2022,LanChemlaKatzir:2023} applied MDL to recurrent neural networks (RNNs) and showed that this allows the networks to find provably correct solutions to a range of tasks that are only approximated using standard methods.

Outside of neural networks, MDL --- and the closely related Bayesian approach to induction --- have been used in a wide range of tasks in which one is required to generalize from very limited data (see \citealp{Horning:1969}, \citealp{Berwick:1982}, \citealp{Stolcke:1994}, \citealp{Grunwald:1996}, and \citealp{Marcken:1996}, and more recently \citealp{RasinKatzir:2016} and \citealp{RasinBergerLanShefiKatzir:2021}. 

\section{\label{sec:mdl:hn}MDL Hopfield Networks}
Implementing MDL within a given formalism involves specifying how to measure both $|G|$, the encoding length of a hypothesis $G$, and $|D:G|$, the encoding length of the training data $D$ given the hypothesis $G$. Moreover, in order to turn MDL into a practical objective function, we should specify how we search the hypothesis space for the choice that optimizes $|G|+|D:G|$. We describe our choices for each of these three decision points below, and discuss more possibilities in section~\ref{sec:disc}. \footnote{The source code used in this paper is available at \url{https://github.com/matanabudy/mdl-hn}.
}

\subsection{Encoding G}
MHN can be described as a set of patterns stored in memory slots. An encoding of these memories is thus an encoding of the `grammar' $G$ (which itself is the basis for an encoding of any data point). To encode such a set of patterns, and get $|G|$ for the MDL objective, one needs to encode the pixel values of these patterns, and concatenate these encodings. Here we used two simple encoding schemes for pixel values in $[0,1]$ (which produced similar results, that we will therefore not distinguish). \textbf{Fixed-Length Encoding:} We encode each pixel value of each pattern with a constant number of bits, say one, resulting in a total encoding length equal to the number of patterns multiplied by the pattern size. \textbf{Prefer-Extreme-Value Encoding:} This encoding favors extreme values: we assign one bit to encode 0 and 1, while the numbers in between are encoded with more bits, with the middle point 0.5 receiving the most complex encoding. For concreteness we chose $[1+ 10 \cdot (value \cdot (1-value))]^{2}$.

\subsection{Encoding D:G}
To encode the data according to the given grammar, we need to specify how each training exemplar can be reconstructed using the grammar (network). For each training exemplar, we first need to encode which memory we should start from to reconstruct it. To do this, we encode the index of the closest memory in the memory slots. We need $\lceil\log_2{\text{num\_memory\_slots}}\rceil$ bits for this encoding.

Next we need to encode the differences between the exemplar and its associated memory. We used the L1-distance between the memory and the training image, multiplied by $\lceil\log_2{\text{pattern\_size}}\rceil$. If all images are bitmaps, then this amounts to encoding where the non-null differences are.

\subsection{Training}
In order to search the hypothesis space for the network that optimizes the MDL objective function we use Simulated Annealing (SA; \citealt{ko83}). Starting from an initial hypothesis, SA repeatedly considers switching to a random neighbor, depending on (a) how the neighbor compares to the current hypothesis, and (b) a temperature parameter. If the neighbor is better (in our case, lower $|G|+|D:G|$), SA switches to it. Otherwise, the worse the neighbor the less likely SA is to switch to it. The temperature parameter determines how risk taking SA is at a given moment: the higher the temperature, the likelier the switch to a worse neighbor. During the search the temperature is slowly lowered, and when it is close to zero the search is effectively greedy.

A random neighbor is derived from the current hypothesis using one of the following operations: 
\begin{inparaenum}
    \item Removing a random memory,
    \item Adding a random training set image to the memories,
    \item Changing an existing memory: Selecting a random memory and applying Gaussian noise to random indices in it,
    \item Crossover: Creating a new memory by averaging two random memories, replacing the originals.    
\end{inparaenum}

\section{\label{sec:exp:1}Experiment 1: The risk with excessive memory in Hopfield Networks}

Much work in the literature has been dedicated to equip Hopfield networks with a large memory capacity. Here we illustrate why the use of these large memory resources should be constrained. 

\paragraph{Data.}
We created a $9\times9$ bitmap of each digit. For each of these `golden' bitmaps, we created several noisy, continuous exemplars of it: each pixel was incremented with Gaussian noise with low or high variance, and the result was clipped to $[0,1]$. Noisy, discrete exemplars were also created, by rounding the continuous exemplars to 0 or 1.
\textbf{Setup.}
HNs were implemented with a version of the HAMUX library \citep{hoover2022a}, updated in \url{https://github.com/bhoov/eqx-hamux}. 

\paragraph{Experimental conditions.}
The different training sets used to train the networks were defined by: 
the number of golden digits represented (1 digit, 2 digits, ..., all 10 digits), 
the number of exemplars per digit (1, 5, 10, 30),
their type (discrete, continuous), their level of noise (low or medium variance). Additionally, the networks were equipped with a memory capacity either corresponding to the cardinality of the training set (unconstrained), or to the cardinality of \emph{golden} digits represented in the training set (golden memories).

\paragraph{Results and discussion.}
Unconstrained MHNs very closely memorized the full input: in the examples inspected, unsurprisingly, memories retrieved from exemplars were diverse, each capturing few exemplars, with distances between an exemplar and its retrieved memory being low, potentially lower than to the underlying golden digit.
On the other hand, with MHNs constrained \emph{a priori} to use no more than the golden number of memories, the learned memories were close to the actual golden digits. (Results became similar to those in Fig.~\ref{fig:avg-memories}). 
MHNs are thus capable of excellent generalization in principle, but only if the memory resources are kept under control. The next experiment shows how to achieve this in a non-supervised manner (i.e.~non-\emph{ad-hoc}), through the MDL objective.

\section{\label{sec:exp:2}Experiment 2: MDL delivers the golden memories (unsupervised)}

\begin{wrapfigure}[14]{r}{5cm}
    \includegraphics[width=5cm, trim=1.5cm 17cm 2cm 0cm, clip]{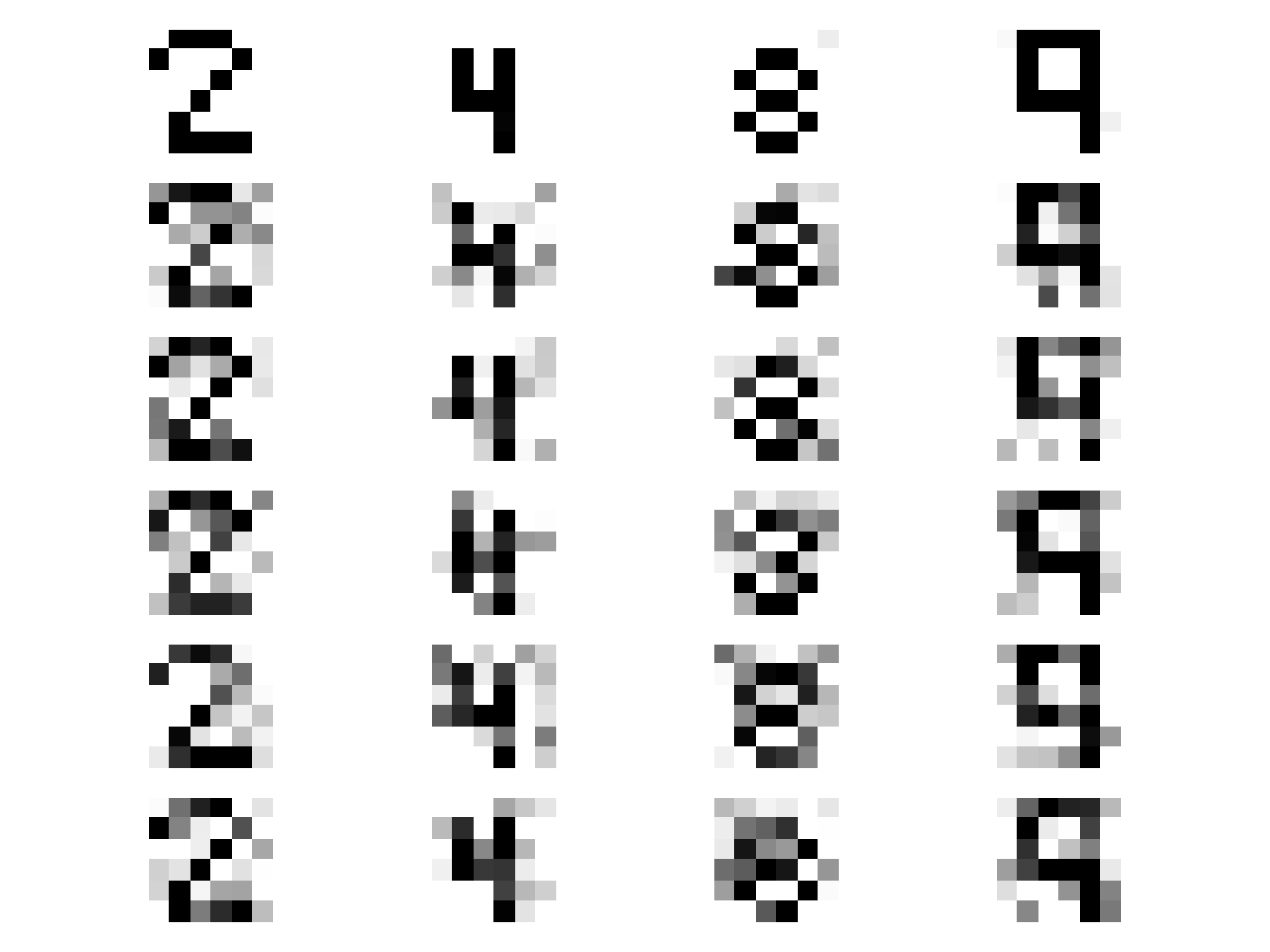}
    \caption{Memory slots learned through MDL training (top row) and exemplars that were matched to each memory.}
    \label{fig:learned-memory-slots}
\end{wrapfigure}

\paragraph{Setup.}
The Minimum Description Length (MDL) principle was incorporated as described in section~\ref{sec:mdl:hn} above to train MHNs, on the same training sets described for Exp.~1.

\begin{wrapfigure}{R}{8cm}
    \includegraphics[width=8cm]{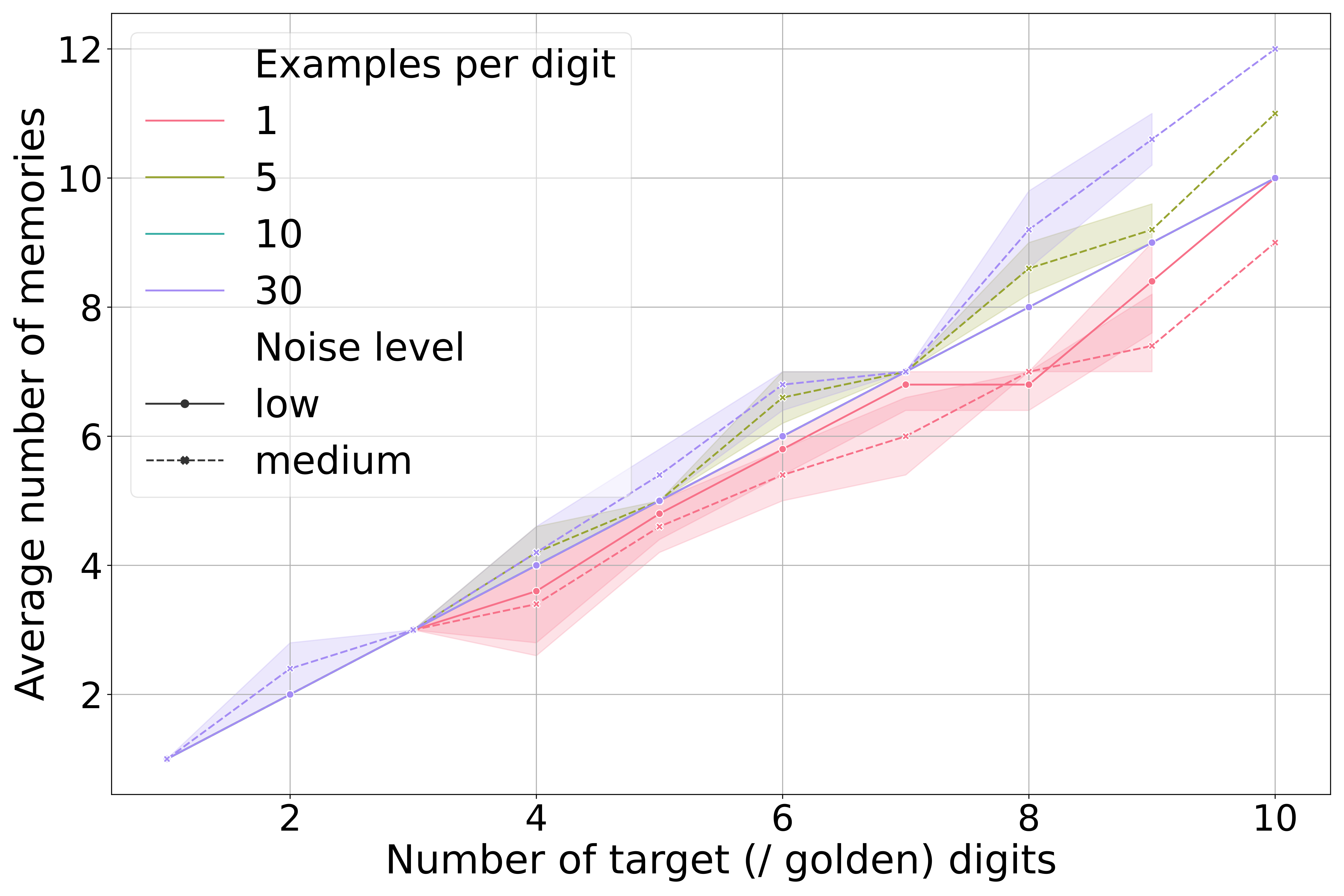}
    \caption{Average number of memory slots found through MDL training on discrete noisy exemplars, compared to the number of golden digits.}
    \label{fig:avg-memories}
\end{wrapfigure}

\paragraph{Results.}
Fig.~\ref{fig:learned-memory-slots} shows the outcome of one simulation. We see that the memories learned through MDL are identical to the target `golden' digits, aside from two grey pixels in digits 8 and 9. This illustrates the two main results. First, the final number of memory slots (which is now adjusted during training), aligns well with the golden number of digits in the training set, see Fig.~\ref{fig:avg-memories} and Fig.~\ref{fig:avg-memories-continuous} in Appx.~\ref{appendix:figs} for continuous noise). This holds strictly true in most cases, except most notably for the high noise conditions, with many exemplars for each digit -- conditions under which an image may be sampled very far from its golden digit, which may reasonably justify the creation of a new category. We come back to these issues in the follow-up below. Second, the learned memories strongly resemble the golden digits: the golden digits were at an average L2 distance of 0.449 for the low noise simulations (1.181 for medium noise) from their associated memories.

\paragraph{Follow-up experiment.}
We conjectured that some deviance from the correct number of memory slots (Fig.~\ref{fig:avg-memories}) may actually be legitimate. A noisy version of a digit could be so far from its source that it would justify creating a new category. As a conservative criterion, we thus consider that an exemplar that ends up closest to a digit that is not its source digit is too noisy. Indeed, it could lead to a new category and an overestimation of categories. Or it could lead to an underestimation, since a golden digit would become redundant as a category if all its exemplars end up closer to another digit.
We therefore re-ran the Exp~.2 with a new dataset, in which we disregarded such exemplars. We found that MDLHNs were even better than before at aligning their number of memories with the (now, more meaningful) golden number, see Fig.~\ref{fig:avg-memories-follow-up} and Fig.~\ref{fig:avg-memories-continuous-follow-up} in Appx.~\ref{appendix:figs}.

\section{\label{sec:disc}Discussion}
We noted that generalization in HNs is limited to the retrieval stage, while storage amounts to rote memorization. We illustrated the problem that this limitation poses when the training data are imperfect, including variants, possibly noisy, of a hidden number of archetypes. In such cases, standard HNs memorize the surface, noisy versions and do not recover the intuitively correct underlying collection of images. We showed how this problem can be remedied by incorporating the generalization principle of MDL into the storage phase: by considering the encoding length of the network (the lower the encoding length, the less capable it is to memorize the data) as a counterweight to the fit to the data, MDLHNs overcame the challenge of generalization and recovered both the correct number of archetypes and the archetypes themselves.

Our proposal is robust to architecture changes. While we illustrated the use of MDL with MHNs, MDL can just as easily be incorporated into classical HNs or other architectures for associative memory. Indeed, as mentioned in section \ref{sec:bg}, MDL has been explored for sequential neural network architectures as well as for entirely different formalisms such as those of formal language theory, and it has also been argued to underlie various mechanisms in human cognition. In this regard, our solution is a principled and very general one to the problem that we noted.

While general, MDL is very tightly connected to representations, which highlights interesting future directions for MDLHNs. Our present illustration relied on simple encoding schemes both for $|G|$ and for $|D:G|$. For both, however, one can consider more sophisticated choices. For instance, in a more realistic setting than ours, the system might have a preference for encoding certain values over others. And for $|D:G|$, our distance function was a convenient choice for the illustration, but it is one that is suitable for static noise that flips random bits. It is not designed to capture, for example, translation, rotation, stretches, reflection, or other operators that are arguably needed for a system that captures human intuitions about visual similarity. Incorporating such operators into $|D:G|$ would be an interesting next step. One simple way to illustrate the relevance of better representations is this. Using our current, naive encodings, a network trained on a single image will have no reason not to memorize that image. With more sophisticated primitives, on the other hand, a reconstruction of an archetype may well be preferable even for a single image. 

Our focus in the present paper has been entirely on the objective function for HNs. For the optimization of the storage phase we simply adopted a general search algorithm (specifically, SA), which supported the recovery of memories in our examples. For setups that are more complex than the one used here, this might prove inadequate, and other optimization algorithms may be more helpful. \citet{LanGeyerChemlaKatzir:2022}, for example, report using a genetic algorithm for optimizing MDL in the domain of RNNs. And other choices might be more helpful still.

\bibliographystyle{unsrtnat}
\setcitestyle{square,numbers,comma}
\bibliography{refs}

\begin{thebibliography}{21}
\providecommand{\natexlab}[1]{#1}
\providecommand{\url}[1]{\texttt{#1}}
\expandafter\ifx\csname urlstyle\endcsname\relax
  \providecommand{\doi}[1]{doi: #1}\else
  \providecommand{\doi}{doi: \begingroup \urlstyle{rm}\Url}\fi

\bibitem[Solomonoff(1964)]{Solomonoff:1964}
Ray~J. Solomonoff.
\newblock A formal theory of inductive inference, parts {I} and {II}.
\newblock \emph{Information and Control}, 7\penalty0 (1 \& 2):\penalty0 1--22, 224--254, 1964.

\bibitem[Rissanen(1978)]{Rissanen:1978}
Jorma Rissanen.
\newblock Modeling by shortest data description.
\newblock \emph{Automatica}, 14:\penalty0 465--471, 1978.

\bibitem[Kolmogorov(1965)]{Kolmogorov:1965}
Andrei~Nikolaevic Kolmogorov.
\newblock Three approaches to the quantitative definition of information.
\newblock \emph{Problems of Information Transmission (Problemy Peredachi Informatsii)}, 1:\penalty0 1--7, 1965.

\bibitem[Chaitin(1966)]{Chaitin:1966}
Gregory~J. Chaitin.
\newblock On the length of programs for computing finite binary sequences.
\newblock \emph{Journal of the {ACM}}, 13:\penalty0 547--569, 1966.

\bibitem[Hinton and Van~Camp(1993)]{hinton_keeping_1993}
Geoffrey~E. Hinton and Drew Van~Camp.
\newblock Keeping the neural networks simple by minimizing the description length of the weights.
\newblock In \emph{Proceedings of the sixth annual conference on {Computational} learning theory}, pages 5--13, 1993.

\bibitem[Zhang and M{\"u}hlenbein(1993)]{ZhangMuhlenbein:1993}
Byoung-Tak Zhang and Heinz M{\"u}hlenbein.
\newblock Evolving optimal neural networks using genetic algorithms with {O}ccam's {R}azor.
\newblock \emph{Complex Systems}, 7\penalty0 (3):\penalty0 199--220, 1993.

\bibitem[Zhang and M{\"u}hlenbein(1995)]{ZhangMuhlenbein:1995}
Byoung-Tak Zhang and Heinz M{\"u}hlenbein.
\newblock Balancing accuracy and parsimony in genetic programming.
\newblock \emph{Evolutionary Computation}, 3\penalty0 (1):\penalty0 17--38, 2020/07/11 1995.

\bibitem[Schmidhuber(1997)]{Schmidhuber:1997}
J{\"u}rgen Schmidhuber.
\newblock Discovering neural nets with low {K}olmogorov complexity and high generalization capability.
\newblock \emph{Neural Networks}, 10\penalty0 (5):\penalty0 857--873, 1997.

\bibitem[Ahmadizar et~al.(2015)Ahmadizar, Soltanian, AkhlaghianTab, and Tsoulos]{AhmadizarSoltanianAkhlaghianTabTsoulos:2015}
Fardin Ahmadizar, Khabat Soltanian, Fardin AkhlaghianTab, and Ioannis Tsoulos.
\newblock Artificial neural network development by means of a novel combination of grammatical evolution and genetic algorithm.
\newblock \emph{Engineering Applications of Artificial Intelligence}, 39:\penalty0 1--13, 2015.

\bibitem[Gaier and Ha(2019)]{GaierHa:2019}
Adam Gaier and David Ha.
\newblock Weight agnostic neural networks.
\newblock \emph{CoRR}, abs/1906.04358, 2019.

\bibitem[Lan et~al.(2022)Lan, Geyer, Chemla, and Katzir]{LanGeyerChemlaKatzir:2022}
Nur Lan, Michal Geyer, Emmanuel Chemla, and Roni Katzir.
\newblock Minimum description length recurrent neural networks.
\newblock \emph{Transactions of the Association for Computational Linguistics}, 10:\penalty0 785--799, 8/2/2022 2022.

\bibitem[Lan et~al.(2023)Lan, Chemla, and Katzir]{LanChemlaKatzir:2023}
Nur Lan, Emmanuel Chemla, and Roni Katzir.
\newblock Benchmarking neural network generalization for grammar induction.
\newblock In \emph{Proceedings of the 2023 CLASP Conference on Learning with Small Data (LSD)}, pages 131--140, Gothenburg, Sweden, September 2023. Association for Computational Linguistics.

\bibitem[Horning(1969)]{Horning:1969}
James Horning.
\newblock \emph{A Study of Grammatical Inference}.
\newblock PhD thesis, Stanford, 1969.

\bibitem[Berwick(1982)]{Berwick:1982}
Robert~C. Berwick.
\newblock \emph{Locality Principles and the Acquisition of Syntactic Knowledge}.
\newblock PhD thesis, MIT, Cambridge, MA, 1982.

\bibitem[Stolcke(1994)]{Stolcke:1994}
Andreas Stolcke.
\newblock \emph{Bayesian Learning of Probabilistic Language Models}.
\newblock PhD thesis, University of California at Berkeley, Berkeley, California, 1994.

\bibitem[Gr\"{u}nwald(1996)]{Grunwald:1996}
Peter Gr\"{u}nwald.
\newblock A minimum description length approach to grammar inference.
\newblock In Stefan Wermter, Ellen Riloff, and Gabriele Scheler, editors, \emph{Connectionist, Statistical and Symbolic Approaches to Learning for Natural Language Processing}, Springer Lecture Notes in Artificial Intelligence, pages 203--216. Springer, 1996.

\bibitem[de~Marcken(1996)]{Marcken:1996}
Carl de~Marcken.
\newblock \emph{Unsupervised Language Acquisition}.
\newblock PhD thesis, MIT, Cambridge, MA, 1996.

\bibitem[Rasin and Katzir(2016)]{RasinKatzir:2016}
Ezer Rasin and Roni Katzir.
\newblock On evaluation metrics in {O}ptimality {T}heory.
\newblock \emph{Linguistic Inquiry}, 47\penalty0 (2):\penalty0 235--282, 2016.

\bibitem[Rasin et~al.(2021)Rasin, Berger, Lan, Shefi, and Katzir]{RasinBergerLanShefiKatzir:2021}
Ezer Rasin, Iddo Berger, Nur Lan, Itamar Shefi, and Roni Katzir.
\newblock Approaching explanatory adequacy in phonology using {M}inimum {D}escription {L}ength.
\newblock \emph{Journal of Language Modelling}, 9\penalty0 (1):\penalty0 17--66, 2021.

\bibitem[Kirkpatrick et~al.(1983)Kirkpatrick, Gelatt, and Vecchi]{ko83}
S.~Kirkpatrick, C.~D. Gelatt, and M.~P. Vecchi.
\newblock Optimization by {Simulated} {Annealing}.
\newblock \emph{Science}, 220\penalty0 (4598):\penalty0 671--680, May 1983.
\newblock ISSN 0036-8075, 1095-9203.
\newblock \doi{10.1126/science.220.4598.671}.
\newblock URL \url{https://www.science.org/doi/10.1126/science.220.4598.671}.

\bibitem[Hoover et~al.(2022)Hoover, Chau, Strobelt, and Krotov]{hoover2022a}
Benjamin Hoover, Duen~Horng Chau, Hendrik Strobelt, and Dmitry Krotov.
\newblock A universal abstraction for hierarchical hopfield networks.
\newblock In \emph{The Symbiosis of Deep Learning and Differential Equations II}, 2022.
\newblock URL \url{https://openreview.net/forum?id=SAv3nhzNWhw}.

\end{thebibliography}

\appendix

\section{Supplementary Figures}
\label{appendix:figs}

\begin{figure}[ht]
    \centering
    \includegraphics[width=0.5\textwidth]{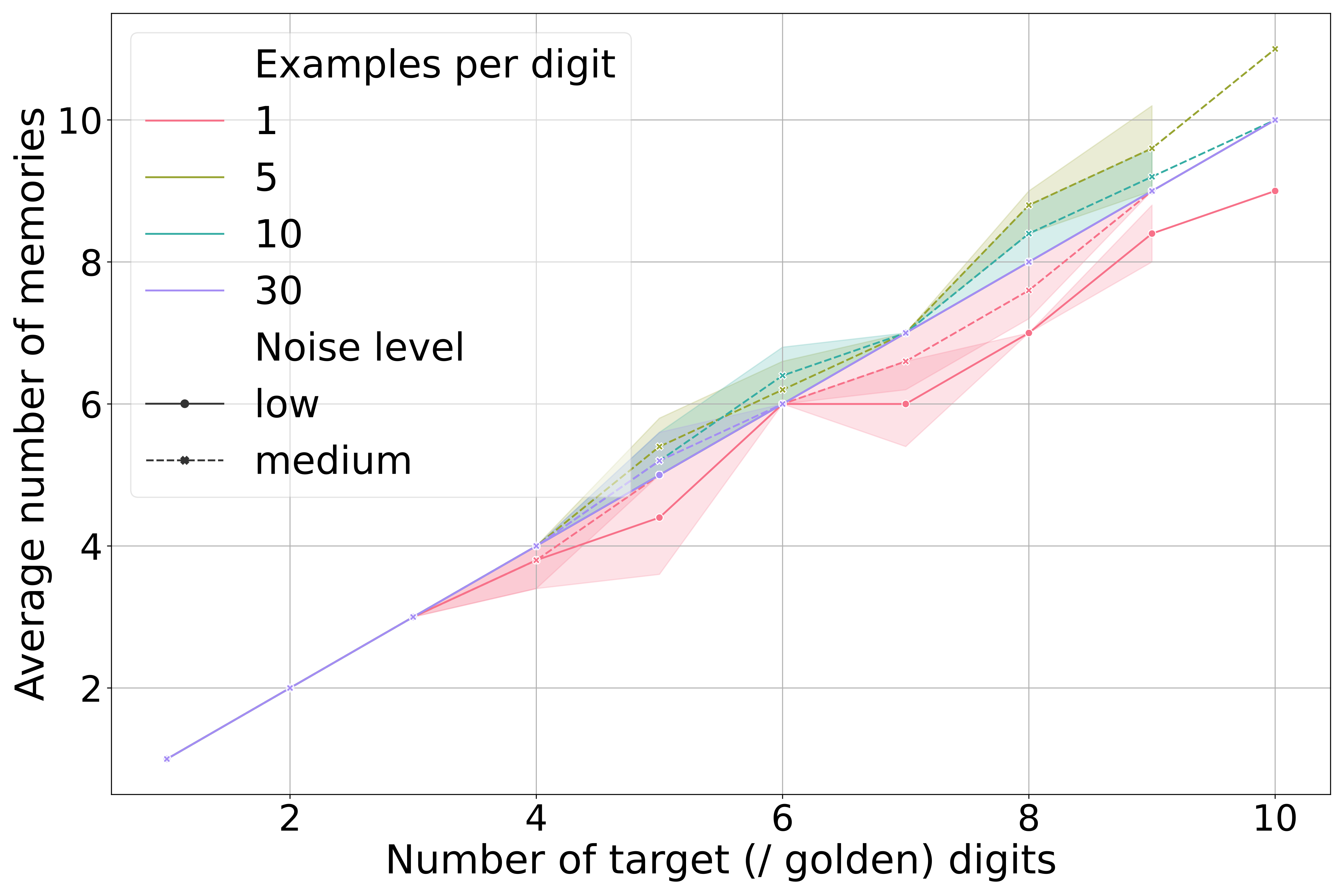}
    \caption{Follow-up to Exp.~1 - average number of memory slots found using MDL, compared to the number of target `golden' digits in the discrete noise training set, while making sure noised exemplars aren't ambiguous between classes.}
    \label{fig:avg-memories-follow-up}
\end{figure}

\begin{figure}[ht]
    \centering
    \includegraphics[width=0.5\textwidth]{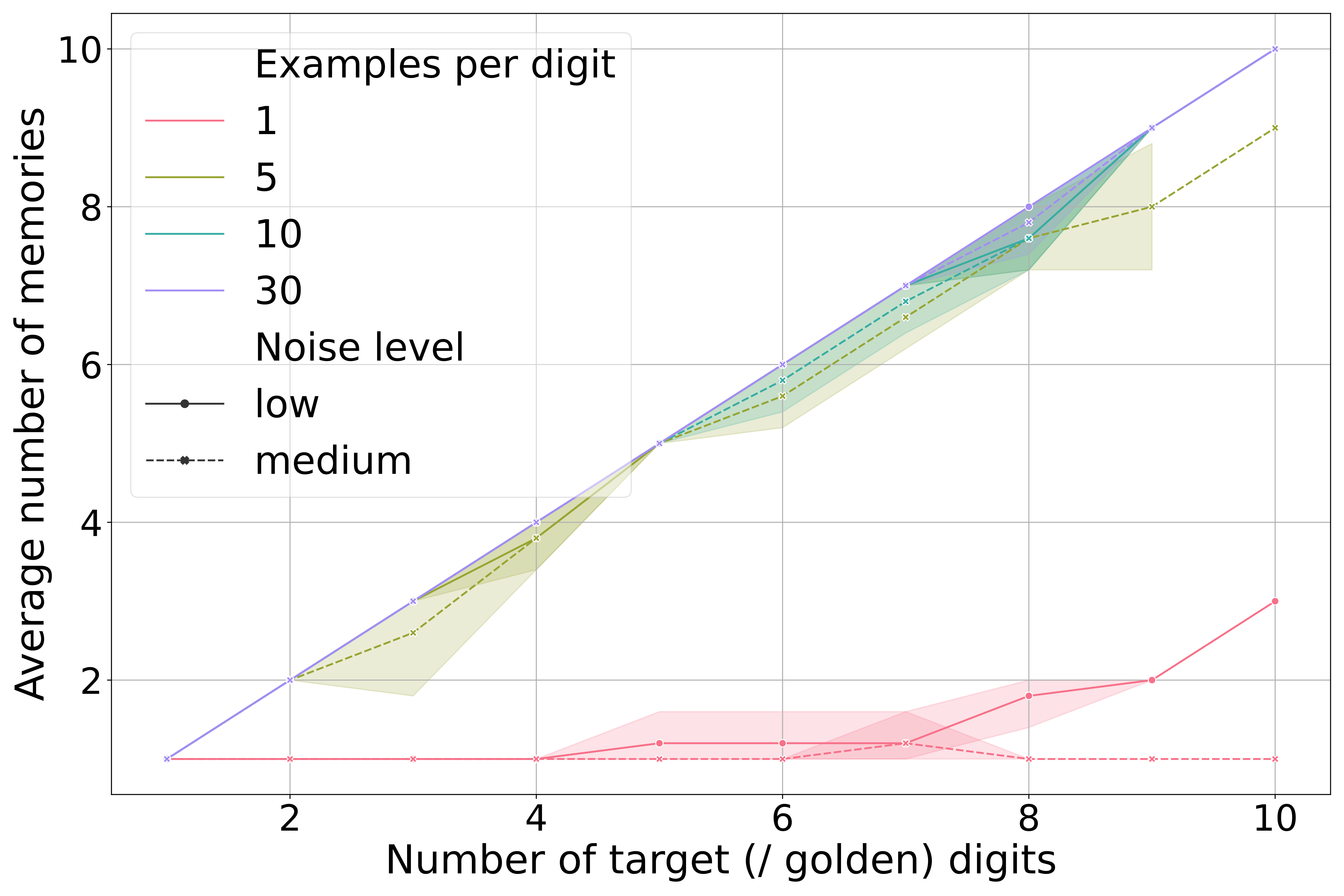}
    \caption{Running Exp.~1 with continuous noise exemplars in the training set}
    \label{fig:avg-memories-continuous}
\end{figure}

\begin{figure}[ht]
    \centering
    \includegraphics[width=0.5\textwidth]{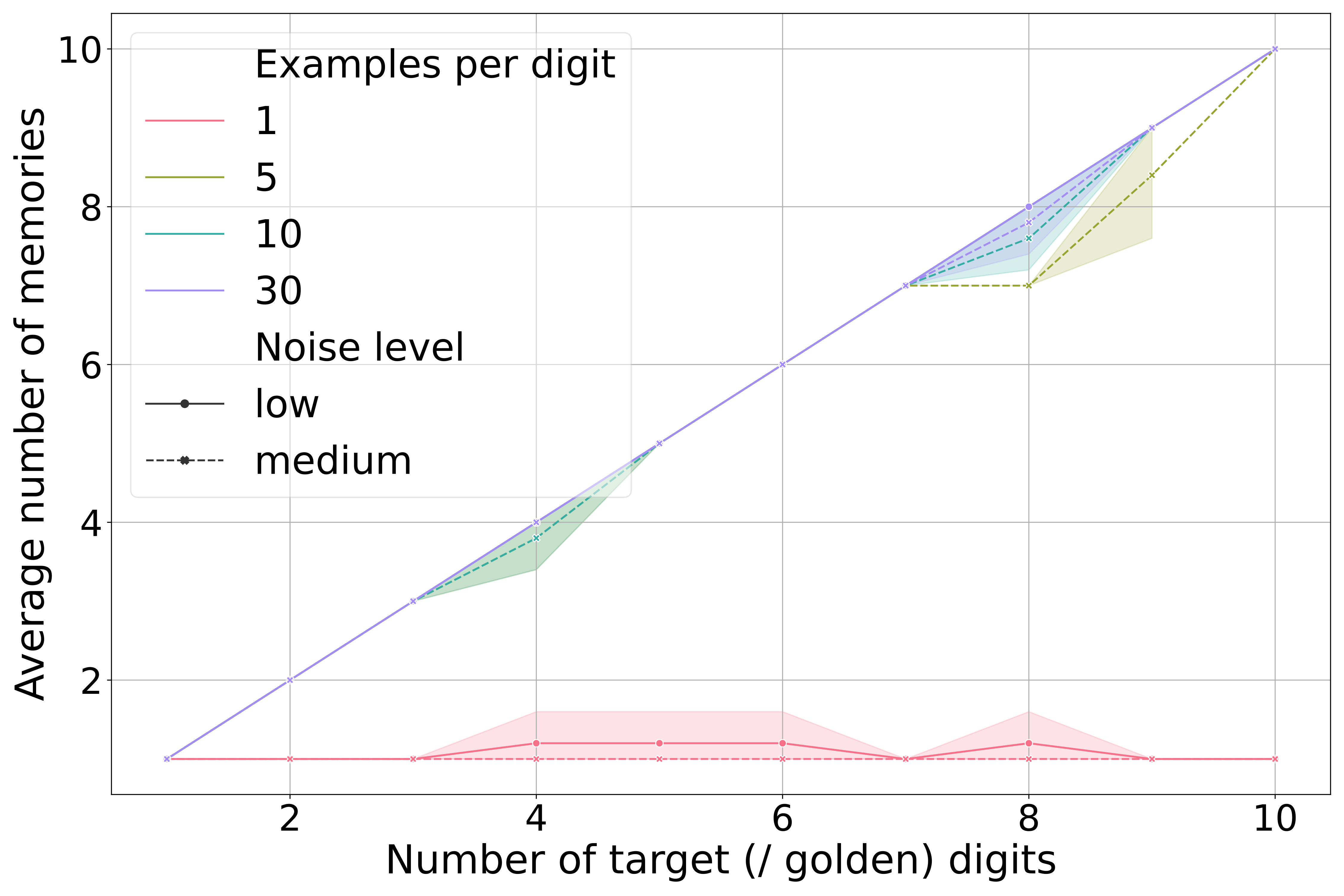}
    \caption{Follow-up experiment for the continuous noise training set}
    \label{fig:avg-memories-continuous-follow-up}
\end{figure}

\end{document}